\pdfoutput=1
\documentclass[11pt]{article}

\usepackage[margin=1in]{geometry}
\usepackage[T1]{fontenc}
\usepackage{lmodern}

\usepackage{amsmath}
\usepackage{amssymb}
\usepackage{pifont}
\newcommand{\cmark}{\ding{51}}
\newcommand{\xmark}{\ding{55}}

\usepackage{graphicx}
\graphicspath{{figs/}}
\usepackage[caption=false,font=normalsize,labelfont=sf,textfont=sf]{subfig}
\usepackage{array}
\usepackage{tabularx}
\newcolumntype{Y}{>{\centering\arraybackslash}X}
\usepackage{multirow}
\usepackage{makecell}
\usepackage{supertabular}
\usepackage{pdflscape}
\usepackage{stfloats}

\usepackage{pgfplots}
\pgfplotsset{compat=1.17}
\usepackage{tikz}
\usetikzlibrary{positioning,shapes,arrows,shadows,patterns}
\usepackage[lined,algonl,algoruled,noend]{algorithm2e}

\usepackage[numbers,sort&compress]{natbib}
\usepackage{textcomp}
\usepackage{verbatim}
\usepackage{enumerate}
\usepackage{setspace}
\usepackage{url}
\usepackage[hidelinks]{hyperref}

\usepackage{authblk}

\title{GERIS: A Game-Theoretic Framework for Filtering Instance-Dependent Label Noise in License Plate Data Augmentation}

\author[1]{Seyedeh~Sara~Jalili~Shani}
\author[2]{Rouhollah~Ahmadian}
\author[2]{Amin~Rahmani}
\author[2]{Mahdi~Bideh}
\author[2,*]{Mehdi~Ghatee}

\affil[1]{Department of Computer Science, University of Alberta\protect\\ Edmonton, Alberta, Canada\protect\\ \texttt{sjalilis@ualberta.ca}}
\affil[2]{Department of Mathematics \& Computer Science, Amirkabir University of Technology\protect\\ Tehran, Iran\protect\\ \texttt{\{rahmadian, amin.rahmani, mahdi.bideh, ghatee\}@aut.ac.ir}}

\date{}

\begin{document}
	
	\maketitle
	{\renewcommand{\thefootnote}{*}\footnotetext{Corresponding author: \texttt{ghatee@aut.ac.ir}}}
	
	\begin{abstract}
		In this paper, we propose GERIS, a game-theoretic framework for instance selection in the data augmentation phase of license plate recognition systems. During augmentation, synthetic license plate images are generated and transformed using stochastic noise to simulate real-world conditions. However, certain noise configurations lead to highly distorted, unreadable images that degrade model performance by introducing instance-dependent label noise. GERIS formulates a non-cooperative game in which each noise vector competes for inclusion in the training set based on its similarity to labeled data and its contribution to model reliability. By identifying and pruning low-quality instances, GERIS improves the overall quality of the augmented dataset. Unlike traditional black-box learning methods, GERIS offers a transparent, theoretically grounded mechanism for data filtering. Experimental results demonstrate that GERIS outperforms existing instance selection methods in terms of classification accuracy and robustness.
	\end{abstract}
	

	\section{Introduction}
	Instance selection is a vital step in data pre-processing for machine learning problems. As datasets become big, the effect of noisy data increases on the training process \cite{nematzadeh2020improving}. Instance selection algorithms aim to mitigate these issues by identifying and removing noisy data from a dataset, resulting in a smaller, more focused training set. Precisely, given a dataset $D$ composed of a training set $T$ and a testing set $U$, let $S \subset T$ be the subset of selected instances resulting from the execution of an instance selection algorithm. Then, the model is trained by $S$, the reduced dataset. Training models on these reduced datasets can be beneficial in several ways. According to the research results of \cite{song2022learning}, the potential benefits of data set filtering and learning include improving the quality of a model, reducing the number of input variables when developing a predictive model, and minimizing empirical error. Identifying detrimental instances and searching for an optimal subset of the training data that minimizes empirical error can also improve the quality of a model. Additionally, data filtering can increase the quality of species records and reduce sample size \cite{van2021impact}. Therefore, dataset filtering can improve model quality by training the model on a reliable dataset.
	
	A license plate recognition (LPR) system is a technology that uses optical character recognition (OCR) to read and recognize license plate numbers from still images or video footage \cite{9929253}. The first  step in a license plate recognition system is detecting and localizing vehicles in an image or a video. There are various techniques to this end. For example, several papers localize vehicle and license plates by object detection models based on CNNs like YOLO \cite{shi2023license}. The next step is to locate the license plate region in the pre-processed image. This step can be done using edge detection, morphological operations, and color segmentation techniques \cite{6717133}. Once the license plate region is located, individual characters must be segmented from the plate. The valuable methods in this step include connected component analysis and contour analysis to extract the characters \cite{fan2022improving,huang2020new}. The final step is to recognize the characters on the license plate using OCR algorithms that match the segmented characters against a database of known characters. The recognition accuracy can be improved using machine learning techniques such as neural networks and deep learning to train the system on a large dataset of license plate images \cite{omar2020cascaded,8698447}. This paper concentrates only on the fourth step to evaluate the proposed model.
	
	Our problem is confined to a specific dataset used for license plate recognition. Since accessing real data from traffic cameras is challenging due to legal barriers, it is reasonable to develop an augmentation framework before model training. Typically, augmentation frameworks introduce various types of noise to the images. However, the synthetic data generated is often unsuitable for training the model directly, as it lacks the realistic variations typically seen in real-world data, such as lighting changes, rotation, shadows, and other environmental factors. Without these features, the generated images contain corrupt examples that resemble instance-dependent noisy labels (IDN) \cite{shorten2019survey}. In this context, instance-dependent noisy labels refer to situations where the noise or distortion introduced during augmentation makes the data instance less reliable for model training, similar to how noisy labels can mislead a model \cite{song2022learning}. Training the model on such noisy instances can lead to inaccurate recognition models \cite{azizian2024preventing}. Thus, it is crucial to generate noisy instances that are more realistic, reflecting the true variability encountered in real-world scenarios, while minimizing the introduction of instance-dependent noise that could impair model performance.
	
	The main challenge surrounding data augmentation is ensuring that the augmented data is representative of the real-world data with which the model will encounter during deployment.
	Suppose the augmented data is distinct from the real-world data.
	In that case, since the model lacks generalization, it will perform poorly when facing new and unseen data. Two common methods are followed to address the above challenge.
	One approach involves applying filtering algorithms to prune synthetic data.
	The second approach includes methods that are based on gradient policy. 
	
	\section{Literature Review}
	
	To provide an overview of the research conducted in the field of data augmentation, this section focuses on two distinct lines of work.
	The first line of research addresses the challenge of augmenting data through the application of filtering algorithms. Various filtering techniques have been explored in different domains. Thus, to maintain relevance to the research objective, this review highlights key algorithms categorized into two main classes: filtering-based instance selection techniques and learning models. While this categorization may become nuanced in some cases, its sole purpose is to provide an overview of counterpart algorithms relevant to our proposed approach. 
	
	Table~\ref{table:comparison} presents a high-level comparison of representative algorithms from each category, offering a concise summary of their characteristics and usage. In the subsequent subsections, we provide a more detailed discussion of these techniques, elaborating on the rationale behind each method, its applicability to noise filtering or instance selection, and its connection to the problem addressed in this work.
	
	\subsection{Filtering-based Instance Selection}
	Filtering-based instance selection algorithms aim to enhance model performance by selecting a subset of instances that are most informative and representative of the underlying data distribution. These methods can be categorized into three main types, each with implications for handling noisy labels, especially those that are instance-dependent.
	
	\subsubsection{Distance-based Filtering Methods}
	Distance-based filtering methods select instances based on their proximity to other instances in the feature space. These methods are effective in identifying and removing redundant or outlier instances, which can help mitigate the impact of noisy labels. Notable algorithms in this category include DBSCAN \cite{song2022learning}, K-nearest Neighbors (KNN) \cite{herrera2022fast}, Isolation Forest \cite{4781136}, K-means \cite{hartigan1979algorithm}, and Condensed Nearest Neighbor (CNN) \cite{hart1968condensed}. By focusing on the spatial relationships between instances, these methods can indirectly address instance-dependent noise by removing instances that are distant from dense regions of the data, which are more likely to be mislabeled.
	
	\subsubsection{Class Distribution Filtering Methods}
	Class distribution filtering methods aim to ensure that the selected instances are representative of the underlying class distribution in the dataset. These methods help preserve the diversity of the dataset and can be particularly useful in identifying and removing instances with noisy labels. Key algorithms in this category include Edited Nearest Neighbor (ENN) \cite{wilson2000reduction}, Neighborhood Cleaning Rule (NCR) \cite{koziarski2020radial}, and Naive Bayes \cite{aridas2019uncertainty}.
	
	\subsubsection{Consistency-based Filtering Methods}
	The third category comprises Consistency-based Filtering methods. These approaches assess the consistency or stability of instances concerning the training data or a learning model. Instances that are more consistent or less susceptible to noise or model variations are considered important and retained, while instances that introduce instability or are less reliable are filtered out. Key algorithms in this class include Cross-Validation \cite{garcia2008extension} and One-Class Support Vector Machine \cite{razzak2020randomized}.
	
	\subsubsection{Noisy Label Detection Methods}
	Label noise falls into two categories: \textit{instance-independent noise (IIN)} (e.g., symmetric/asymmetric noise) and \textit{instance-dependent noise (IDN)}, where errors correlate with features, posing unique generalization challenges \cite{song2022learning,zhu2024label}. 
	\textbf{Sample selection} methods exploit the memorization effect, filtering unreliable data using distance-based criteria (e.g., MentorNet \cite{jiang2018mentornet}) or neighborhood consistency checks (e.g., ConFrag \cite{kim2024sample}). \textbf{Label correction} techniques instead modify annotations via prediction confidence \cite{chen2023two} or clean subset agreement \cite{zheng2021meta}, while \textbf{hybrid methods} combine both approaches \cite{chang2023csot}. 
	Alternative solutions include robust loss functions \cite{liu2020peer}, regularization (e.g., adaptive checkpointing \cite{azizian2024preventing}), meta-learning for sample reweighting \cite{wang2020training}, and contrastive learning \cite{yao2021jo}, though the latter may conflate semantic similarity with noise.

	\subsection{Learning Models}
	Learning models are typically employed to perform classifications and predict patterns, which, in our context, can assist in classifying redundant transformation functions and removing them from our instances. Some applicable learning models in our setting are Multilayer Perceptron (MLP) and Support Vector Machines (SVM).
	
	\subsection{Policy-gradient Methods}
	The second line of research falls under policy-gradient methods which focus on determining the optimal combination and parameters for data augmentation transformations. These studies aim to identify the most effective and efficient ways to transform the existing data, enhancing its diversity and generalization capabilities. 
	\\
	Regarding this line of work, the principal augmentation method to discuss is AutoAugment.
	In contrast to traditional manual data augmentation methods, AutoAugment is an automated procedure that searches for improved data augmentation policies. One limitation of AutoAugment is that it requires powerful computational resources  \cite{xu2023comprehensive}.
	Several versions of AutoAugment have been proposed as an enhancement of this data augmentation method such as RandAugment, AugMix, TrivialAugment, and DeepAugment \cite{mumuni2022data}.
	The remainder of this paper is structured as follows: Initially, we will give a brief description of the data generating framework. Then the underlying concepts of game theory will be introduced. Next, we will
	give an overall explanation of the applied game to further elaborate on how we derive an ideal instance of
	data with it. Finally, we will test the algorithm via two evaluation methods and compare our method with its counterparts.
	
	To clearly differentiate our work from existing techniques, we have conducted an extensive performance comparison, as outlined in Section \ref{sec:experimental_results}.  However, through a comprehensive analysis, our work stands out from its counterparts in several key aspects, which are highlighted below:
	\begin{itemize}
		\item \textbf{Novel Methodology:} We are the first to apply a game theory approach to dataset pruning, which sets our work apart from other filtering techniques. By leveraging game theory principles, we introduce a novel methodology that optimizes the dataset pruning process, resulting in superior accuracy compared to alternative approaches.
		\item \textbf{Computational Efficiency:} Our approach exhibits lower computational cost compared to gradient policy-based methods. Unlike those methods that require training a given model in each iteration, our approach streamlines the computational requirements, enabling faster and more efficient dataset pruning without sacrificing accuracy.
		\item \textbf{Dimensionality Reduction:} We have significantly reduced the dimensionality of the problem by leveraging noise vectors as the primary input and eliminating the reliance on complete images. This strategic decision was made to mitigate the challenges associated with the curse of dimensionality. By reducing the dimensionality, our approach achieves improved efficiency and scalability while maintaining the effectiveness of dataset pruning.
	\end{itemize}
	
	\newpage
	\begin{table*}
		\centering
		\footnotesize
		\renewcommand{\arraystretch}{1.1}\setlength{\extrarowheight}{1.5pt}
		\caption{A review of algorithms used for instance selection.}
		\label{table:comparison}
		\begin{tabular}{|>{\centering\arraybackslash}m{0.025\linewidth}|>{\raggedright\arraybackslash}m{0.16\linewidth}|>{\centering\arraybackslash}m{0.14\linewidth}|>{\raggedright\arraybackslash}m{0.47\linewidth}|>{\centering\arraybackslash}m{0.07\linewidth}|}
			\hline
			\multicolumn{3}{|c|}{\textbf{Instance Selection Methods}} & \multicolumn{1}{c|}{\textbf{Description}} & \multicolumn{1}{c|}{\textbf{Ref.}} \\
			\hline
			\multirow{11}{*}{\hspace{0pt}\rotatebox{90}{Filtering-based Instance Selection}} & \multirow{5}{1.0\linewidth}{\hspace{0pt}Distance-based filtering selects instances by their proximity to others.}                                                                                                                                                                                                                                                                & DBSCAN                                 & A density-based clustering algorithm groups data points without requiring a preset cluster count, selecting representatives from dense regions. & \cite{song2022learning}                                                                               \\ 
			\cline{3-5}
			&                                                                                                                                                                                                                                                                                                                                                                                                                                    & K-nearest Neighbors (KNN)              & KNN classifies data points by their nearest neighbors' majority class. K-means clusters data by similarity, selecting centroids as representatives. & \cite{herrera2022fast}                                                                        \\ 
			\cline{3-5}
			&                                                                                                                                                                                                                                                                                                                                                                                                                                    & Isolation Forest                       & An isolation forest detects anomalies by isolating rare, distinct instances using random forests. & \cite{4781136}                                                                                    \\ 
			\cline{3-5}
			&                                                                                                                                                                                                                                                                                                                                                                                                                                    & K-means                                & K-means clusters data into \textbf{k} groups by minimizing squared distances to centroids, using them as representatives. & \cite{hartigan1979algorithm}                                                                      \\ 
			\cline{3-5}
			&                                                                                                                                                                                                                                                                                                                                                                                                                                    & Condensed Nearest Neighbor       & A method that iteratively selects only correctly classified, essential instances for an accurate reduced dataset. & \cite{hart1968condensed}                                                                          \\ 
			\cline{2-5}
			& \multirow{4}{1.0\linewidth}{\hspace{0pt}Class distribution filtering selects instances to represent the dataset's class diversity.}                                                                       & Edited Nearest Neighbor (ENN)          & A method that removes misclassified or redundant instances by comparing them to their nearest neighbors, boosting model performance.                                                                                                                                           & \cite{wilson2000reduction}                                                                        \\ 
			\cline{3-5}
			&                                                                                                                                                                                                                                                                                                                                                                                                                                    & Neighborhood Cleaning Rule (NCR)       & An ENN extension that removes Tomek links---mutual nearest neighbors of different classes---to enhance class separability. & \cite{koziarski2020radial}                                                                        \\ 
			\cline{3-5}
			&                                                                                                                                                                                                                                                                                                                                                                                                                                    & Naive Bayes                            & A probabilistic classifier that estimates the probability of an instance belonging to each class based on the class distribution in the training data.                                                                                                                                                                                                                   & \cite{aridas2019uncertainty}                                                                      \\ 
			\cline{3-5}
			&                                                                                                                                                                                                                                                                                                                                                                                                                                    & Uncertainty Quantification             & Measures labeling uncertainty by modeling noisy vs. true label distributions, improving noise robustness across domains and models. & \cite{rottmann2023automated,ahmadian2024uncertainty}                                                                     \\ 
			\cline{2-5}
			& \multirow{2}{1.0\linewidth}{\hspace{0pt}Consistency-based filtering keeps instances that are stable and reliable, removing noisy or inconsistent ones.} & Cross-Validation (CV)                  & Evaluate the stability or consistency of instances by performing repeated cross-validation and removing instances that show high variance in their performance across different folds.                                                                                                                                                                                   & \cite{garcia2008extension}                                                                        \\ 
			\cline{3-5}
			&                                                                                                                                                                                                                                                                                                                                                                                                                                    & One-Class Support Vector Machine & A machine learning algorithm that learns a decision boundary around a set of instances, identifying regions where most instances belong, and enabling the detection of outliers or instances that deviate from the majority.                                                                                                                                             & \cite{razzak2020randomized}                                                           \\ 
			\hline
			\multirow{3}{*}{\hspace{0pt}\rotatebox{90}{Learning Models}}                     & \multirow{3}{1.0\linewidth}{\hspace{0pt}They classify and predict patterns, helping identify and remove redundant transformation functions.}                                                                                                                                                                                             & Multilayer Perceptron (MLP)            & A feedforward artificial neural network model consisting of multiple layers of interconnected nodes (neurons) that learn to classify or predict patterns by adjusting the weights associated with each connection.                                                                                                                                                       & \cite{rumelhart1986learning}                                                                      \\ 
			\cline{3-5}
			&                                                                                                                                                                                                                                                                                                                                                                                                                                    & Support Vector Machines (SVM)          & Supervised machine learning algorithm that finds an optimal hyperplane to separate data points of different classes by maximizing the margin between them, achieving high classification accuracy.                                                                                                                                                                       & \cite{roy2019fuzzy}                                                                          \\ 
			\cline{3-5}
			&                                                                                                                                                                                                                                                                                                                                                                                                                                    & Fusion Function                          & This approach focuses on noise filtering using fusion functions, such as majority voting and averaging filters. It requires that a label is agreed upon by the majority of classifiers. Experiments conducted on noisy datasets have shown that majority vote filters are more effective in detecting incorrectly labeled instances compared to consensus filters or individual classifiers. This method significantly enhances learning performance. & \citep{ahmadian2025enhancing,ahmadian2024improved,wahmadian2023driver}                                                                     \\
			\hline
		\end{tabular}
	\end{table*}
	
	\newpage
	\section{Proposed methodology}
	\label{sec:proposed_methodology}
	This section consists of two parts. Firstly, the framework for generating artificial license plates is described. To this end, the various noises applied to the simple plates are listed. Then, the game of pruning corrupt generated instances is informed. This part explains the underlined theory and the method which with the proposed game is evaluated.
	\subsection{Data generator framework}
	Generating license plates needs a straightforward framework. Thus, we propose a data generation framework with different steps that help us produce an authentic replica of an actual license plate image. We entitled our framework as ''Synthetic license plate data generator via a non-cooperative game'' or shortly GERIS.
	\begin{figure}[ht]
		\centering
		\includegraphics[width=1.0\textwidth]{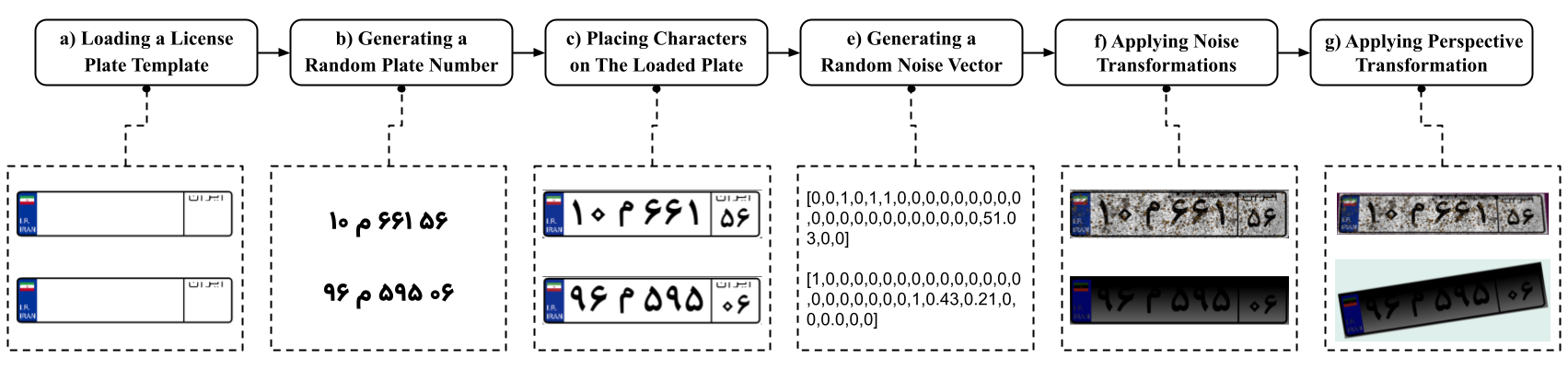}
		\caption{The architecture of generating license plate.}
		\label{fig:generator}
	\end{figure}
	
	\begin{table*}[http]
		\footnotesize
		\centering
		\renewcommand{\arraystretch}{1.3}
		\caption{Transformation Functions}
		\label{table:noises}
		\begin{tabular}{|c|p{12.8cm}|}
			\hline                                                                                                                                	\textbf{Transformation} & \multicolumn{1}{c|}{\textbf{Description}}                        \\ \hline
			Salt and Pepper                  & The Salt and Pepper noise is a simple noise that simulates dead pixels in a camera and digital noise. This noise uses two values corresponding to salt and pepper, respectively, $v_s$ and $v_p$.                                                                                                                                                                                                                                                                             \\ \hline
			Blur                             & The Blur noise uses Gaussian and median filters to add a blur effect to the images. This noise uses two parameters, filter size, and variance. An increase in filter size or variance will increase the blur.                                                                                                                                                                                                                                                                                                       \\ \hline
			Circular Light                   & The circular light noise is the simulation of the reflection of light in some images in the form of fuzzy circles.
			To obtain such noise, first, we apply a median filter to smooth it out and prevent unwanted circular lights. After that, we convert the image format from RGB to YUV and add noise to its Y channel. The noise is a Gaussian kernel amplified by a factor $l_p$.                                                                                                               \\ \hline
			Light                            & The light noise is the replica of high exposure in images that adds extra brightness to an image. This phenomenon makes text to be less readable. To construct such noise, we proceed as the Circular light noise and convert the image format to YUV after applying median smoothing. Following these steps, we add light noise as a constant to the Y channel of the image.                                                                                                                                                                                                                                                                                                        \\ \hline
			Gradient Light                  & The gradient light noise is another light noise that can produce noises similar to light beams. This noise occurs when a light source is in the environment and the camera's lens is not clean. The initial steps for producing this noise are similar to other light noises. After obtaining YUV channels, we select a linear path and add a constant to the Y channel. In the last step, we convert the image to RGB channels and obtain a noisy image.                                                                                                                                                                                                                                                                                                                                                                                                                        \\ \hline
			Negative                         & The negative noise works the opposite way of the light noises. This noise lowers the exposure and light in the image by deducting a constant from the Y channel of the YUV format of the image. YUV format of the image and reverse conversions are similar to light noises, and the only difference is the subtraction part.                                                                                                                                                                                                                                                                                                                                           \\ \hline
			Mixture                          & Since a license plate can have multiple corrupting noises simultaneously, these types of noisy plates must be included in the dataset.
			In order to generate such images, we apply some of our previously defined noises to an image and obtain a mixture of noises, as seen in Figure \ref{fig:generator}.
			However, these mixtures can cause a plate to become unreadable to both humans and machines. Thus, further procedures are required to avoid such data samples.                                                                                                                                                                                                                                                                         \\ \hline
			Perspective                      & The angle that a camera has toward a scene can cause perspective. Perspective makes objects in an image have different widths and heights from their actual dimensions. To mimic this phenomenon, we apply a perspective transformation to an image. Our transformation uses four control points. We also consider four directions in each control point and randomly select a direction for each of the points to obtain a transformation. \\ \hline
		\end{tabular}
	\end{table*}
	
	First, we load some background images, called templates, that act as the plate, then we randomly generate characters of the license plate and place them on the loaded template. For further details, we add different non-perspective and perspective noises that help replicate a detected license plate in an actual image.
	Figure \ref{fig:generator} illustrates the proposed framework steps.
	
	Each license plate consists of different layers. The first layer is the template layer that acts as the metal plate used in an actual license plate. To obtain realistic templates, we investigated all the different license plates installed on vehicles. Then we carefully designed each template using additional software and, as a result, produced 13 different templates equal to the number of all categories of license plates in Iran.
	
	We use two general categories of noises. The two categories are computed noises and overlay noises.
	The first category contains noises that can be computed using
	a formula in real-time, e.g., salt and pepper noise.
	In Table \ref{table:noises}, we described the computed noises.
	In the noise vectors, illustrated in Fig \ref{fig:generator}.e, the 10th to 30th components are associated with the first category of noises.
	The noises in the second category are in the form of transparent images, e.g., scratch marks.
	In the noise vectors, illustrated in Fig \ref{fig:generator}.e, the 0th to 9th components are associated with the second category of noises.
	
	\subsection{Instance selection}
	This section explains how game theory is applied for pruning the corrupt instances generated by the framework.
	The utilized game is a non-cooperative game in which the players, i.e. noise vectors, have to choose a strategy between \textit{correct} or \textit{corrupt}.
	The proposed labeling algorithm consists of a game played between an unlabeled data point and its neighbors.
	The payoff for each player is calculated based on its Mahalanobis distance from its neighboring vectors.
	In the remainder of this section, we will thoroughly explain how this technique enables labeling the whole dataset by manually labeling a small fraction.
	
	\subsubsection{Game theory}
	In general, game theory is known to study the interaction of decision-makers in situations of conflict or cooperation.
	A game is a description of strategic interaction that includes at least two players, each of which poses a set of possible strategies. The strategies that the players end up picking will determine the outcome of the game. Each possible outcome obtained by playing a set of strategies establishes the payoffs for the players. These payoffs indicate the players' gain of the reached outcome \cite{straffin1993game}.
	The game involves mutual influence among participants, as each participant's decisions affect the choices of others. Its overall objective is for every participant to strategically balance their own interests while maximizing collective gains, assuming full rationality.
	Game theory allows us to model such phenomena in a game. It provides us with the analytical tools to find possible solutions for them. Thus, its use has become increasingly prevalent in solving problems in economics, biology, and politics.
	
	\subsubsection{Non-cooperative games}
	The game we attempt to describe is based on the game proposed in \cite{almogahed2015neater}, which is called NEATER, for balancing the dataset. However, by applying some changes in the original game, this study tends to prune the corrupt instances generated in the data augmentation process. The proposed game is classified as a non-cooperative game. We first introduce the notions of such games and then represent how we modeled our problem into one. The further brought terminology is highly similar to what is used in \cite{nash1996non} and is close to the notions used in \cite{almogahed2015neater} to further facilitate the appliance of their method in our context. A normal form game consists of a set of players $i \in N$, in which the set $N=\{1,2,\cdots ,n\}$ is finite.
	Each player $i$ chooses a strategy from the pure strategy space $S_i={1,2,\cdots ,k_i }$, which is the set of pure strategies available to player $i$. Moreover, we have a collection of payoff functions $u_i$, each of which maps an $n$-tuple known as the pure strategy profile $s=(s_1,s_2,\cdots ,s_n)$, $s_i \in S_i$ being the pure strategy for player $i$, into the player $i$'s utility. This means that a player's payoff depends both on their and their opponent's strategies. The opponents of player $i$ are referred to as $-i$.
	In a normal-form game, players play only to maximize their utility based on a detailed description of the moves and information available to them. In a non-cooperative game, these self-interests pursuing individuals cannot cooperate or form coalitions. 
	Another important class of strategies in a game is mixed strategies. A mixed strategy $x_i$ for any player $i$ is a probability distribution over their pure strategies, i.e., $S_i$. This randomization process for each player takes place independently of others. A mixed strategy $x_i$ can be represented by a vector with its $d$-th component being the probability of assigning the pure strategy $d\in S_i$   to player $i$. The dimension of this vector is the number of possible pure strategies that the player can choose. Note that all the probabilities are positive, and they sum to one: $\sum_{j=1}^{\left | S_i  \right |}x_{ij}=1$ for all $i\in N$.
	Assuming the vector $x=(x_1,x_2,\cdots ,x_N)$ is the mixed strategy profile with each $x_i$ being the mixed strategy followed by the player $i$, the payoffs to a profile of mixed strategies are the expected values of the corresponding pure-strategy payoffs. Suppose $S$ is the pure strategy space for each pure strategy profile $s$ and $s \in S$. The expected value of the payoff to player $i$ with the mixed strategy $x_i$ is given by: $u_i(x) = \sum_{s \in S} x(s) u_i(s)$.
	Remember that $u_i (s)$ is the payoff player $i$ obtains if all the agents follow the pure strategy profile $s$. Furthermore, $x(s)$ is the probability of choosing pure strategy profile $s$, i.e. the product of probabilities or $\prod_{i=1}^{N} x_j(s_j)$.
	
	A pure strategy $s_i$ for the $i$-th player can also be represented as a mixed strategy vector. If $s_i=d$ and assuming that $d \in S_i$, then we can show $s_i$ by assigning 1 to $ x_i$'s $d$-the component and zero to the rest of the components. From now on, we will denote this representation by $e_i^d$. Another important notion to introduce is $(x_i,x_{-i})$. The term $x_{-i}$ is the mixed strategy profile $x$ without its $i$-th component. Thus, the tuple $(x_i,x_{-i})$ indicates adding $x_i$ back to $x_{-i}$. This notion is used while defining Nash equilibrium as we want to compare the utility of player $i$ with different sets of mixed strategy vectors. 
	According to \cite{nash1996non}, a mixed strategy profile $x^*$ is a Nash equilibrium, if for all players $i$, $u_i(x_i^*,x_{-i}^*) \geq u_i (x'_i,x_{-i}^*)$ for all possible $x'_i$ in the mixed strategy space. The same definition holds for pure strategies. Nash equilibrium conveys that a game reaches equilibrium if all the players act by a strategy profile in which each player's strategy is an optimal response to the other players' strategies. A game might have a more complex strategy profile that results in a Nash equilibrium. However, in a normal-form game with finite players and a finite set of actions or strategies, we will at least have a Nash equilibrium of mixed strategies.
	
	\begin{figure}[ht]
		\centering
		\includegraphics[width=0.90\textwidth]{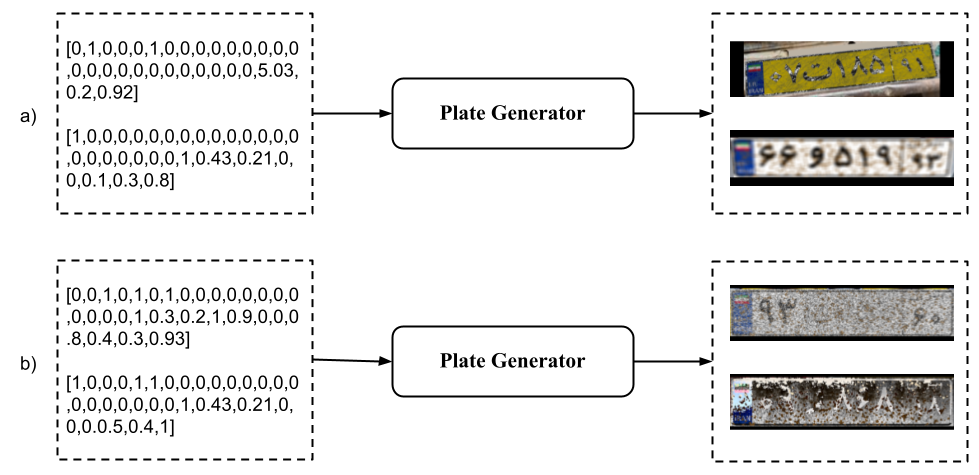}
		\caption{Four samples of correct and corrupt noise vectors. (a) refers to the correct samples, and (b) refers to the corrupt samples.}
		\label{fig:NoiseVectorToImage}
	\end{figure}
	
	\subsubsection{Modeling the problem into a normal form game}
	Consider having a dataset consisting of noise vectors. This dataset is formed of two classes of noise vectors dubbed \textit{correct} and \textit{corrupt}. The \textit{correct} noise vectors, as shown in Figure \ref{fig:NoiseVectorToImage}.a, when added to the image of a license plate, will result in a recognizable yet more natural image. On the other hand, the \textit{corrupt} noise vectors, as shown in Figure \ref{fig:NoiseVectorToImage}.b, will ruin the image and make it unrecognizable for the model.
	Thus, having a dataset of noise vectors $D$, we have a set of \textit{correct} vectors $D_c$ and a set of \textit{corrupt} vectors $D_p$.
	The corrupt noise vectors must be removed from the dataset since their inclusion leads to training an inaccurate model.
	However, labeling these noise vectors is quite challenging when we have an extremely large dataset. To tackle this problem, we suggest manually labeling a random minor subset of the dataset, with the help of which we can classify the rest of the dataset as \textit{correct} or \textit{corrupt} using the proposed game.
	This way, we can label the whole dataset by only labeling a tiny fraction of it. 
	The labeling algorithm is formulated as a non-cooperative normal-form game. The game is symmetric with two players, and given that its purpose is establishing the label of a noise vector, it is played among an unlabeled vector and a group of its neighbors with size $h$. To determine the neighborhood set we use a k-nearest neighbors approach based on Mahalanobis distance. For each unlabeled noise vector, the 
	$h$ closest vectors in the labeled and unlabeled sets are selected. Suppose that the set of all neighboring vectors of an unlabeled data point is shown by the set $R$. This set can comprise both labeled vectors, denoted by the $R_l$, and unlabeled ones, denoted by $R_u$. Note that $R_l \cup R_u = R$. The game occurs between the unlabeled data and one neighbor at a time. The strategy space for each player $i$ is to be either labeled as \textit{corrupt} or \textit{correct} or in other words, $S_i=\{p,c\}$ ($p$ stands for corruPt and $c$ for correCt). The payoff for each player is the sum of the payoffs they receive by playing each of the two-player games with their neighbors.
	Considering that a matrix can represent the payoff function of a game and there are two actions available for each player, each game played between players $i$ and $j$ will have a two-by-two matrix $A_{ij}$ as its payoff function. The matrix $A_{ij}$ is formulated as $A_{ij}=m_{ij}×I_2$, where $m_{ij}$ is the inverse of the Mahalanobis distance between the noise vectors $i$ and $j$. Hence, for a mixed strategy profile $x=(x_1,x_2,\cdots,x_h)$ the utility of player $i$ can be derived from: $u_i(x) = \sum_{j=1}^{h} x_i^T A_{ij} x_j$.
	\newline
	As mentioned before, the neighboring vectors of unlabeled noise vectors can either belong to $R_l$ or $R_u$.
	Since the vectors in $R_l$ are already labeled, they can only play the pure strategies $p$ or $c$. These pure strategies are shown by $e_i^c= e_i^1=(1,0)$ and $e_i^p= e_i^2=(0,1)$. With $R$ being the opponents of player $i$ in the game, the above summation can be written as: 
	\begin{align}
		u_i(x) = \sum_{j \in R_u} x_i^T A_{ij} x_j + \sum_{d=1}^{2} \sum_{j \in R_l} x_i^T A_{ij} e_j^d
	\end{align}
	
	Having formulated the problem set as a game, all that is left to do is compute the outcome of the game that results in classifying all the unlabeled noise vectors.
	
	\subsubsection{Replicator dynamics}
	Replicator dynamics is an evolutionary selection model used in game theory to study how the frequency of different strategies in a population changes over time, and can help to calculate the Nash equilibrium of a symmetric game \cite{weibull1997evolutionary}.
	In replicator dynamics, the population is represented by a distribution of strategies, where each strategy has a particular frequency or proportion in the population. The system's dynamics are governed by the fitness of each strategy, which measures how well it performs relative to other strategies in the population. The fitness of a strategy is typically defined in terms of the payoff it receives in a game against other strategies. Since the game we described in the previous section has a finite set of players and strategies and, more importantly, is symmetric, meaning players earn the same payoff when making the same choice against similar choices of their competitors, we can use replicator dynamics as a tool to reach its Nash equilibrium. 
	According to \cite{weibull1997evolutionary}, replicator dynamics presumes that individuals can only be programmed to pure strategies. So in such case, a mixed strategy $x$ is regarded as the population state with each $x_i$ demonstrating the share of individuals in our population that are following the pure strategy $i$. In modeling the population's evolution, its state must change over time to demonstrate the supremacy of a strategy over others.
	Since in our environment, the population was non-overlapping, we could use the discrete-time model with each period $t=0,1,2,\cdots$  representing a generation rather than its continuous-time counterpart. Non-overlapping populations refer to a situation where the individuals in a population are divided into separate, distinct groups that do not interact or compete with each other.
	Each group represents a separate sub-population, and the dynamics of each sub-population is modeled independently of the others. In the discrete-time replicator dynamics, $x(t)=(x_1,x_2,\cdots,x_k)$ with $k$ being the number of available strategies, is the $t$-th generation's distribution over the available pure strategies. Now suppose that payoffs represent the number of offspring each sub-population will have in the next generation and $p_i$ ($t$) is the number of individuals of type $i$ in a population. Evidently, the size of population is $p(t)=\sum_{i=1}^k p_i(t)>0$. The size of the sub-population playing strategy $i$ in the next generation relies on the size of the sub-population in the current generation and the number of offspring it is allowed. Hence, as shown in \cite{weibull1997evolutionary} the dynamics is obtained by:
	\begin{align}
		p_i (t+1)=[\alpha + u(e^i,x(t))] ~ p_i (t)
	\end{align}
	where $\alpha$ is the birthrate, a parameter that regulates the growth or decay rate of strategies. Large values of $\alpha$ will lead to slower convergence, and smaller values give faster but less stable convergence. Moreover, $u_i(e^i,x(t))$ is the expected payoff of playing the pure strategy $i$ when the population is in state $x(t)$.
	Consider that in our game, since the utility is computed by summing all the payoffs of two-person games played among neighbors, the population state $x(t)$  must include the population states of all the neighbors. Thus, to avoid confusion, we reformulate the population state as $\bar{x}(t)=(x_1 (t),x_2 (t),\cdots,x_h (t))$. Then we will have:
	\begin{align}
		p_i^d (t+1)=[\alpha + u_i (e_i^d  ,\bar{x}_{-i} (t))] p_i^d (t)
	\end{align}
	where $i$ is the player in a set $R$ and $d \in \{c,p\}$. 
	Now summing over all of the pure strategies will leave us with the following:
	\begin{align}
		p_i (t+1)=[\alpha+ u_i (\bar{x}(t))] p_i (t)
	\end{align}
	where $u_i (\bar{x}(t))=\sum_{d=1}^2 x_i^d (t) u_i (e_i^d ,\bar{x}_{-i} (t))$.
	In replicator dynamics, the probability of playing a strategy is equivalent to the proportion of individuals with that strategy in the population.
	Accordingly, $x_i^d (t)$ or the probability of playing the strategy $d$, is also determined by: $x_i^d (t)=  (p_i^d (t))/(p_i (t))$.
	Now in order to calculate the frequency of a strategy for a play $i$, for instance, the $c$ or correct strategy, we have to divide the two above equations, which gives us the following:
	\begin{align}
		x_i^c (t+1)=\frac{\alpha+ u_i (e_i^c  ,\bar{x}_{-i} (t))}{\alpha+ u_i (\bar{x}(t))} x_i^c (t)
	\end{align}
	More precisely, we have:
	\begin{align}
		x_i^c (t+1)=\frac{\alpha+ \sum_{j=1}^{h} e_i^{c^T}A_{ij}x_j}{\alpha+ \sum_{j=1}^{h} x_i^T(t)A_{ij}x_j(t)} x_i^c (t)
	\end{align}
	
	Taking into consideration that we have two strategies, thus two frequencies, the probability of playing the $p$ or \textit{corrupt} strategy is calculated by: $x_i^p (t+1)= 1- x_i^c (t+1)$.
	The above dynamics will lead to the growth of sub-populations associated with better-than-average strategies and the decline of worse-than-average strategies. In the long run, this process will converge to a Nash equilibrium \cite{cressman2014replicator}.
	
	Eventually, after enough execution of the above algorithm, the probability of belonging to a specific class for each unlabeled data point can be used for labeling these data points. This means that after reaching convergence, the label associated with a higher probability will be assigned to the unlabeled noise vector. For a more explicit description of the GERIS, one can refer to Algorithm \ref{algo}.
	
	\begin{algorithm}[H]
		\label{algo}
		
		\SetCommentSty{emph}
		\SetKwInOut{Input}{input}
		\caption{{\footnotesize Labeling noise vectors with GERIS}} 
		\Input{A set $S$ of random noise vectors, $1/n$ of which are labeled.}
		\BlankLine
		\For{each unlabeled noise vector $i$ in $S$}{
			\BlankLine
			\tcp{set initial probabilities of each label to $0.5$}
			\emph{$x_i^p, x_i^c \leftarrow 0.5, 0.5$}
			\BlankLine
		}
		\BlankLine
		\tcp{{\scriptsize probabilities of labeled vectors are given in accordance with their labels}}
		\BlankLine
		\While{convergence condition is not reached}{
			\BlankLine
			\For{each unlabeled noise vector $i$ in $S$}{
				\BlankLine
				\tcp{{\scriptsize $u_c$ and $u_p$ are the utility of pure strategies $c$ and $p$}}
				\tcp{{\scriptsize $u_t$ is the utility of the whole population }}
				\emph{$u_c, u_p, u_t \leftarrow 0, 0, 0$}\;
				\emph{$R \leftarrow h$ number of $i$'s neighbors}
				\BlankLine
				\For{each neighbor $j$ in $H$}{
					\BlankLine
					\emph{$u_{c/p} \leftarrow u_{c/p} + e_i^{{c/p}^T}A_{ij}x_j$}\;
					\emph{$u_t \leftarrow u_t + x_i^TA_{ij}x_j$}\;
				}
				\BlankLine
				\emph{$x_i^{c/p} \leftarrow \frac{\alpha + u_{c/p}}{\alpha + u_t}$}\;
				\BlankLine
			}
		}
		\BlankLine
		\For{each unlabeled vector $i$ in $S$}{
			\BlankLine
			\If{$x_i^c$ is bigger than $0.5$}{
				\emph{label $i$ as $c$}\;
			}
			\Else{\emph{label $i$ as $p$}\;}
		}
	\end{algorithm}
	
	\subsection{Evaluation methodology}
\begin{figure}[ht]
		\centering
		\includegraphics[width=0.75\textwidth]{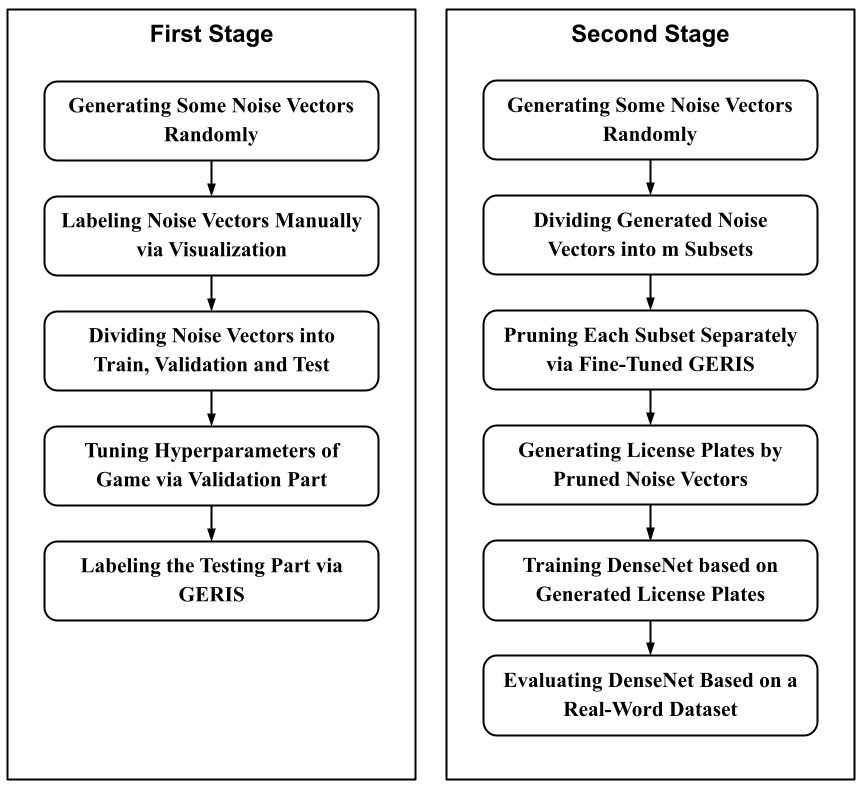}
		\caption{Two stages of evaluation methodology.}
		\label{fig:EvaluationLicensePlateGame}
	\end{figure}

	The proposed model's performance is evaluated through two stages. In the first stage, a direct evaluation was carried out by comparing manually labeled data to the labels achieved by our algorithm. The second stage aims to measure the accuracy of an Image-To-Text model that was trained with the generated license plates after being filtered by our algorithm. Figure \ref{fig:EvaluationLicensePlateGame} shows the evaluation pipeline of the proposed game.
	
	\subsubsection{Measurement}
	The proposed model was evaluated by F-measure. This is due to the fact that the portions of correct and corrupt data are significantly different, and the dataset is imbalanced. Thus, using metrics like accuracy will not provide us with a precise demonstration of the model's performance in classifying the minority class. 
	Furthermore, based on the types of noises incorporated in a noise vector, two classes of corruption could take place.
	The first class is the \textit{corrupt perspective} and the second class is the \textit{corrupt non-perspective}.
	This division was crucial due to the distinct nature of these noise types. For instance, rotating a license plate that contains a \textit{6}, might transform it to an image with a \textit{9}.
	In this case and other similar situations, the \textit{perspective noises} might lead to the corruption of an image while the addition of other non-perspective noises could maintain the image's readability.
	The opposite of the above example is also probable.
	For instance, a generated image of a license plate with the proper perspective could be impossible to read because of its
	blurriness.
	Therefore, each noise vector has two labels, one is chosen between \textit{correct perspective} and \textit{corrupt perspective}, and the other one can either be \textit{correct non-perspective} or \textit{corrupt non-perspective}.
	In addition, since the number of corrupt noise vectors being formed of perspective noises is not equal to the number of those being formed of non-perspective noises, the F-measure is calculated for each class separately. Otherwise, the evaluation of the model's performance at detecting the two mentioned corruptions will be ambiguous.
	As a result, the following equations are used for evaluating the pruning mechanism:
	\begin{align}
		F1 &= (F1_{p} + F1_{np})/2
		\\
		Precision &= (Precision_{np} + Precision_{p})/2
		\\
		Recall &= (Recall_{p} + Recall_{np})/2
	\end{align}
	where the notation of $np$ refers to the F-measure of labeling vectors with non-perspective noises and $p$ refers to the F-measure of labeling perspective noises.
	Finally, the label of the $i$-th noise vector is calculated by the following equation:
	\begin{align}
		Label^{(i)} = Label_{p}^{(i)} * Label_{np}^{(i)}
	\end{align}
	
	A noise vector is labeled as corrupt if either one of the perspective or non-perspective labels is determined as corrupt.
	
	\subsubsection{First stage methodology}
	In this approach, a small fraction of generated noise vectors are labeled manually.
	To this end, we generate license plates via the framework, and add random noise vectors to it.
	If after adding each noise vector the quality of the image is maintained, the correct label is assigned to that vector. Subsequently, the labeled noise vectors are divided into three sets called train, validation, and test.
	The optimal values for hyperparameters of the game are obtained via a grid search using the validation set.
	These values are also employed for pruning license plates in the second stage.
	The train part of the dataset is given as a labeled sample for GERIS in order to predict labels of the data points in the test set.
	As labels of the test set are known previously, the computed F-measures can be evidence of the model's performance.
	
	\subsubsection{Second stage methodology}
	This stage is more important as compared with the previous stage, since alongside indicating the model's performance it is also in accordance with our main purpose, i.e., training the recognition model.
	Initially, a large number of noise vectors are generated randomly.
	Then, the noise vectors are divided into $m$ subsets, as labeling all of them in one process requires large memory usage.
	Therefore, GERIS is utilized with the previously computed hyperparameters to label the noise vectors in $m$ chunks.
	Next, the noise vectors labeled as corrupt are removed, and license plates are generated based on the correct noise vectors. Since DenseNet \cite{huang2017densely} is a well-known model for classifying characters, it was chosen and trained by our generated license plates.
	DenseNet's performance can prove the reliability of applying GERIS to filter the dataset.
	For this aim, a real-world license plate dataset is then used to evaluate DenseNet.

	\section{Experimental results}
	\label{sec:experimental_results}
	In order to create a dataset that accurately reflects inputs of traffic cameras, our framework produces a noise vector by combining one to three randomly selected noise types from Table \ref{table:noises}.
	This is because the experiments found that incorporating more than three noise types ensures the corruption of an image.
	In addition, a grid search for finding the optimal hyperparameters for the proposed game revealed the following optimal values: $h = 40$, $\alpha = 0.999999$, and 20 iterations for convergence.
	
	Experiments were conducted on an Apple MacBook Pro (M1 Pro chip, 16GB RAM) running macOS Monterey 12.6. The code utilized Python 3.9.2 with PyTorch 2.0.0 (compiled for Metal GPU acceleration), scikit-learn 1.2.2, and NumPy 1.26.4. PyTorch leveraged the M1 GPU via the Metal Performance Shaders (MPS) backend for hardware-accelerated training.
	
	\subsection{Sensivity Analysis of Training Data Ratio}
\begin{figure}[http]
		\centering
		\footnotesize
		
		\begin{tikzpicture}
			\pgfplotsset{width=0.7\textwidth, height=6.5cm,compat=1.9}
			\begin{axis}[
				xlabel={Ratio of training data (\%)},
				ylabel={Mean F1 score (\%)},
				xmin=10, xmax=80,
				ymin=50, ymax=100,
				enlargelimits=true,
				legend pos=south west,
				ymajorgrids=true,
				grid style=dashed,
				legend columns = 5,
				legend style={
					at={(0.5,-0.3)}, 
					anchor=north,
					legend cell align=left,
				},
				major y tick style = transparent,
				major x tick style = transparent,
				]
				\legend{$Recall_{corrupt}$, $Recall_{correct}$, $F1$, $Precision_{corrupt}$, $Precision_{correct}$}
				\addplot[line width=0.5mm,color=orange,mark=otimes*,]
				coordinates {(80,87.86)(70,87.02)(60,87.27)(50,85.48)(40,82.96)(30,80.56)(20,76.46)(10,69.67)};
				\addplot[line width=0.5mm,color=orange,mark=diamond*,]
				coordinates {(80,92.51)(70,92.95)(60,93.00)(50,93.40)(40,93.64)(30,94.11)(20,94.78)(10,95.80)};
				\addplot[line width=0.5mm,color=green,mark=triangle*,]
				coordinates {(80,79.66)(70,79.62)(60,79.66)(50,79.70)(40,79.45)(30,79.45)(20,79.42)(10,79.02)};
				\addplot[line width=0.5mm,color=cyan,mark=otimes*,]
				coordinates {(80,57.08)(70,57.58)(60,57.35)(50,57.52)(40,57.32)(30,57.21)(20,57.46)(10,57.36)};
				\addplot[line width=0.5mm,color=cyan,mark=diamond*,]
				coordinates {(80,99.18)(70,99.04)(60,99.11)(50,99.02)(40,98.88)(30,98.81)(20,98.62)(10,98.33)};
				
			\end{axis}
		\end{tikzpicture}
		\caption{Analysis of training size ratio in terms of mean F1 score.}
		\label{fig:anlysis_ratio}
	\end{figure}
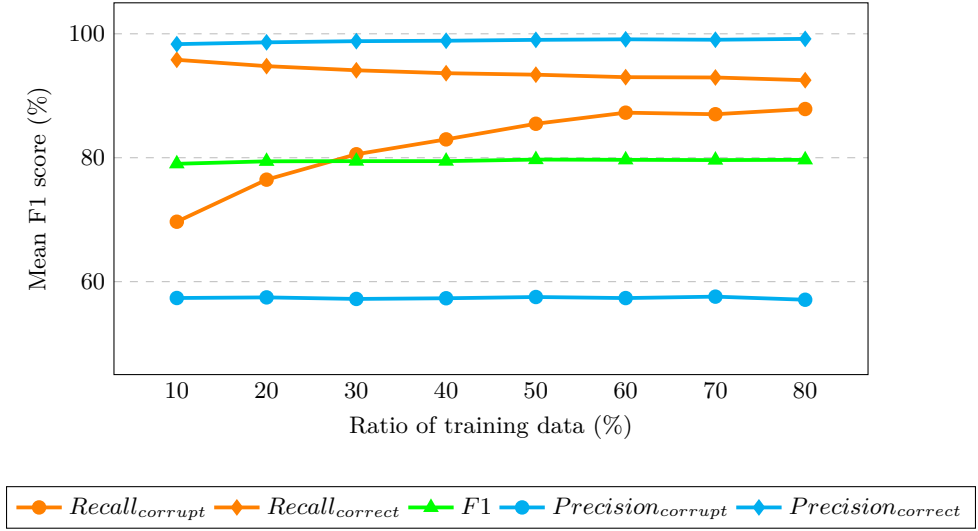

	Regarding the algorithm's sensitivity to the ratio of the train set to the test set, Figure \ref{fig:anlysis_ratio} shows the mean F1 score computed for several arbitrary ratios based on the test set given in the first stage.
	The means calculated from the F1 scores were obtained from 100 experiments in which the testing data was picked randomly.
	The figure demonstrates that as the proportion of manually labeled noise vectors decreases from 80\% down to 10\%, the overall mean F$_1$‐score of the GERIS framework remains remarkably stable around 79\%, indicating a high degree of robustness to reduced labeling effort. While the recall for corrupt instances ($Recall_{\text{corrupt}}$) drops modestly from approximately 87.9\% at 80\% labeling to 69.7\% at 10\%, and $Precision_{\text{correct}}$ declines slightly from 99.2\% to 98.3\%, the complementary metrics ($Recall_{\text{correct}}$ and $Precision_{\text{corrupt}}$) actually improve or remain flat as labeling is reduced. This suggests that GERIS can maintain balanced performance even when only a small fraction of noise vectors are labeled. From a scalability standpoint, requiring as little as 10\% labeled data yields near‐optimal F$_1$ performance, substantially lowering manual annotation cost without sacrificing model effectiveness, and thus supports deployment on larger datasets with limited labeling budgets.
	
	\subsection{Experimental results of the first stage}
	In the final step of this stage, we compared the proposed method with several well-known classification algorithms, given that GERIS, too, performs a classification task over noise vectors. To evaluate the performance of the proposed scheme, 2000 random noise vectors were generated. Our benchmarks included classical machine learning methods such as KNN, MLP, Naive Bayes, SVM, and Random Forest, as well as two modern deep learning baselines: Transformer-based models and Squeeze-and-Excitation (SE) networks. Transformer architectures leverage self-attention mechanisms to capture long-range dependencies and global feature interactions, making them particularly effective for learning complex data relationships \cite{vaswani2017attention}. In contrast, SE networks introduce channel-wise attention through a two-step process of “squeeze” (global information aggregation) and “excitation” (adaptive feature recalibration), allowing the model to emphasize more informative features while suppressing noise \cite{hu2018squeeze}. While Transformers are adept at modeling global context, SE networks focus on refining local discriminative features—both offering complementary advantages in representation learning. In the context of tabular noise vectors, Transformers help uncover high-level correlations, whereas SE modules enhance signal-to-noise discrimination.
	
	Table \ref{table:benchmark} shows the benchmark results.
	Each experiment was repeated 100 times, and the average result is reported in the table.
	It is crucial to note that $Precision_{corrupt}$ and $Recall_{corrupt}$ are of more importance, as keeping a corrupt noise vector in the dataset is more detrimental than removing a correct noise vector.
	As evident in Table \ref{table:benchmark}, the F-measures of GERIS are higher than those of other stated classification algorithms.
	Three critical findings emerge from Tables~\ref{table:benchmark} and~\ref{table:resource_usage}:
	\begin{itemize}
		\item \textbf{Detection Superiority}:
		GERIS achieves \textbf{85.84\%} corrupt recall ($\pm0.002$) --- \textbf{31.16 pp higher} than SE Networks ($54.68\% \pm0.003$), \textbf{33.04 pp higher} than Transformers ($52.80\% \pm0.002$), and \textbf{1.57 pp higher than Naïve Bayes ($84.27\% \pm0.002$)}. This comes while maintaining near-perfect correct label precision (\textbf{99.04\% vs. Naïve Bayes' 99.15\%}), reducing false positives by \textbf{2.1$\times$} compared to SE Networks.
		
		\item \textbf{Stability-Accuracy Tradeoff}: 
		While Naïve Bayes shows competitive corrupt recall (84.27\%), its \textbf{corrupt precision is 55.07\% --- 2.28 pp lower than GERIS (57.35\%)}. The SE Network's high corrupt precision ($62.01 \pm0.018$) proves brittle, suffering \textbf{31.16 pp lower recall} than GERIS. Transformers show similar limitations, with precision-recall imbalance ($61.72$ vs. $52.80$).

		\item \textbf{Operational Efficiency}: 
		GERIS runs \textbf{1.15$\times$ faster} than Naïve Bayes ($2.66$s vs. $3.07$s) while achieving \textbf{1.17 pp higher F1 ($79.66$ vs. $78.49$)}. Compared to deep learning methods, GERIS is \textbf{4.9$\times$ faster} than SE Networks ($12.99$s) and \textbf{12.9$\times$ faster} than Transformers ($34.35$s).
	\end{itemize}
	The results demonstrate that GERIS successfully balances:
	\begin{itemize}
		\item \textbf{Effectiveness}: Outperforms all baselines in F1, including \textbf{+1.17 pp over Naïve Bayes ($79.66$ vs. $78.49$)}
		\item \textbf{Reliability}: \textbf{1.2$\times$ lower F1 variance} than Naïve Bayes ($\pm0.013$ vs. $\pm0.015$)
		\item \textbf{Safety}: \textbf{99.04\% correct precision} vs. Naïve Bayes' \textbf{99.15\%} (marginally lower by 0.11 pp but with \textbf{1.57 pp higher corrupt recall})
		\item \textbf{Speed}: Processes 377 samples/second vs. Transformers' 29 samples/second.
	\end{itemize}
	
	This combination proves particularly effective for noise filtering in class-imbalanced scenarios ($29.8\%$ corrupt labels). While Naïve Bayes achieves marginally higher correct precision ($99.15\%$), GERIS provides better balance with \textbf{1.57 pp higher corrupt recall} and \textbf{1.15$\times$ faster execution}. The explicit pattern storage architecture, while memory-intensive ($1289$MB), enables auditability absent in both traditional statistical methods and modern neural approaches.
	
	\begin{table*}[http]
		\centering
		\caption{Benchmark of Filtering Methods (Mean $\pm$ Std over 100 Trials)}
		\renewcommand{\arraystretch}{1.3}
		\label{table:benchmark}
		\resizebox{\textwidth}{!}{%
			\begin{tabular}{|c|c|c|c|c|c|}
				\hline
				\textbf{Method} & $\textbf{F1}$                & $\mathrm{\textbf{Precision}}_{\mathrm{\textbf{corrupt}}}$       & $\textbf{Precision}_{\textbf{correct}}$ & $\textbf{Recall}_{\textbf{corrupt}}$            & $\textbf{Recall}_{\textbf{correct}}$ \\ \hline
				KNN             & 73.61 ($\pm$ 0.002)          & 48.62 ($\pm$ 0.000)          & 97.72 ($\pm$ 0.000)    & 48.69 ($\pm$ 0.000)            & 99.48 ($\pm$ 0.0) \\ \hline
				MLP             & 73.28 ($\pm$ 0.003)          & 49.54 ($\pm$ 0.000)          & 97.04 ($\pm$ 0.000)    & 46.85 ($\pm$ 0.000)            & \textbf{99.84 ($\pm$ 0.000)} \\ \hline
				Naïve Bayes & 78.49 ($\pm$ 0.015)          & 55.07 ($\pm$ 0.000)          & \textbf{99.15 ($\pm$ 0.000)}    & 84.27 ($\pm$ 0.002)          & 92.11 ($\pm$ 0.0) \\ \hline
				SVM             & 73.67 ($\pm$ 0.003)          & 49.16 ($\pm$ 0.000)          & 97.61 ($\pm$ 0.000)    & 48.27 ($\pm$ 0.000)            & 99.69 ($\pm$ 0.000) \\ \hline
				Rand Forest & 73.80 ($\pm$ 0.002)           & 49.05 ($\pm$ 0.000)          & 97.76 ($\pm$ 0.000)    & 48.78 ($\pm$ 0.000)            & 99.64 ($\pm$ 0.000) \\ \hline
				One Class SVM & 56.45 ($\pm$ 0.023)           & 29.60 ($\pm$ 0.002)          & 87.06 ($\pm$ 0.001)    & 39.63 ($\pm$ 0.003)            & 87.75 ($\pm$ 0.001) \\ \hline
				Isolation Forest & 60.49 ($\pm$ 0.016)           & 37.09 ($\pm$ 0.001)          & 88.71 ($\pm$ 0.001)    & 44.98 ($\pm$ 0.003)            & 88.87 ($\pm$ 0.001) \\ \hline			
				KMeans & 39.61 ($\pm$ 0.168)           & 20.86 ($\pm$ 0.021)          & 81.41 ($\pm$ 0.016)    & 53.17 ($\pm$ 0.038)            & 49.46 ($\pm$ 0.064) \\ \hline
				SE Network       & 77.27 ($\pm$ 0.035)  & \textbf{62.01 ($\pm$ 0.018)}                                       & 97.97 ($\pm$ 0.000)                     & 54.68 ($\pm$ 0.003)                  & 99.02 ($\pm$ 0.000)                  \\ \hline
				Transformer      & 76.45 ($\pm$ 0.033)  & 61.72 ($\pm$ 0.023)                                       & 97.88 ($\pm$ 0.000)                     & 52.80 ($\pm$ 0.002)                  & 99.21 ($\pm$ 0.000)                  \\ \hline
				\textbf{GERIS}             & \textbf{79.66 ($\pm$ 0.013)} & 57.35 ($\pm$ 0.000) & 99.04 ($\pm$ 0.000)    & \textbf{85.84 ($\pm$ 0.002)} & 93.20 ($\pm$ 0.000)    \\ \hline
			\end{tabular}
		}
	\end{table*}
	
	\begin{table*}[http]
		\centering
		\renewcommand{\arraystretch}{1.3}
		\caption{Resource Usage Profile of Filtering Methods (Mean $\pm$ Std over 100 Trials)}
		\label{table:resource_usage}
		\begin{tabular}{|c|c|c|}
			\hline
			\textbf{Method}  & \multicolumn{1}{c|}{\textbf{Run Time (s)}}          & \multicolumn{1}{c|}{\textbf{Memory Usage (MB)}}                    \\ \hline
			KNN              & 2.09 ($\pm$ 0.58)                      & 212.38 ($\pm$ 18.29)                                \\ \hline
			MLP              & 6.18 ($\pm$ 1.58)                      & 194.34 ($\pm$ 26.57)                                \\ \hline
			Naïve Bayes      & 3.07 ($\pm$ 0.05)                      & 183.33 ($\pm$ 5.65)                                 \\ \hline
			SVM              & 2.13 ($\pm$ 0.05)                      & 210.71 ($\pm$ 8.67)                                 \\ \hline
			Rand Forest      & 1.80 ($\pm$ 0.02)                      & 241.83 ($\pm$ 9.57)                                 \\ \hline
			One Class SVM    & 2.17 ($\pm$ 0.06)                      & 203.37 ($\pm$ 21.27)                                \\ \hline
			Isolation Forest & 1.29 ($\pm$ 0.03)                      & 215.09 ($\pm$ 3.88)                                 \\ \hline
			KMeans           & 1.96 ($\pm$ 0.60)                      & 224.98 ($\pm$ 4.00)                                 \\ \hline
			SE Network      & 12.99 ($\pm$ 1.20)                      & 186.80 ($\pm$ 16.93)                                \\ \hline
			Transformer      & 34.35 ($\pm$ 0.48)                      & 234.89 ($\pm$ 69.66)                                \\ \hline
			GERIS            & \multicolumn{1}{c|}{2.66 ($\pm$ 0.05)} & \multicolumn{1}{c|}{1289.59 ($\pm$ 84.08)} \\ \hline
		\end{tabular}
	\end{table*}
	
	\subsection{Experimental results of the second stage}
	In this stage, a real-world license plate dataset is used as a touchstone to assess the model's performance in character recognition.
	Doing so, 50000 random noise vectors are produced by the previously introduced generator to train the model with different input datasets.
	Table \ref{table:resultsModel} demonstrates the result of the experiments.
	In these experiments, six approaches were pursued.
	\begin{enumerate}
		\item A character recognition process takes place based on an unfiltered real input dataset.
		\item The model is trained with the unfiltered data produced by the framework.
		\item The same dataset in the previous approach was used, but it was filtered with Naive Bayes.
		\item Both real and augmented data, which was filtered by Naive Bayes, were involved in the training of the character recognition model.
		\item The data generated by the framework and filtered by GERIS was given as an input to the model.
		\item Both real and augmented data, which was filtered by GERIS, were involved in the training of the character recognition model.
	\end{enumerate}
	
	To conduct these experiments, the dataset is divided into two subsets of digits and alphabets.
	We split the dataset to carry out the detection of each group of characters with their own specialized models.
	This division was easily done due to the fact that in a license plate, alphabets always appear once and are located in a fixed position.
	Accordingly, the two DenseNet models, one for digits and the other for alphabets are trained and evaluated separately.
	The illustrated data in Table \ref{table:resultsModel} proves that using the introduced framework to augment data along with GERIS as a filtering method causes a significant improvement in the overall performance of the model. We specifically compared the impact of our method to that of Naive Bayes, as it achieved the highest performance among the benchmark models presented in Table \ref{table:benchmark}.
	As seen in the obtained results, regarding the dataset of digits, the accuracy, F1 score, precision, and recall of the baseline model were increased by 10.05\%, 17\%, 17.85\%, and 15.32\% respectively. Regarding the dataset of alphabets, an increase of 26.78\%, 31.09\%, 26.24\%, and 33.13\% was seen for the same mentioned measures respectively.
	
	\begin{table*}[http]
		\centering
		\renewcommand{\arraystretch}{1.2}
		\caption{Benchmark of Character Recognition}
		\label{table:resultsModel}
		\resizebox{\textwidth}{!}{%
			\begin{tabular}{|cc|c|cccc|cccc|}
				\hline
				\multicolumn{2}{|c|}{\textbf{Training Data}}                  & \multirow{2}{*}{\textbf{\begin{tabular}[c]{@{}c@{}}Filtering\\ Method\end{tabular}}} & \multicolumn{4}{c|}{\textbf{Digits}}                                                                                                     & \multicolumn{4}{c|}{\textbf{Alphabets}}                                                                                                  \\ \cline{1-2} \cline{4-11} 
				\multicolumn{1}{|c|}{\textbf{Real Data}} & \textbf{Synthetic} &                                                                                      & \multicolumn{1}{c|}{\textbf{Accuracy}} & \multicolumn{1}{c|}{\textbf{F1}}    & \multicolumn{1}{c|}{\textbf{Precision}} & \textbf{Recall} & \multicolumn{1}{c|}{\textbf{Accuracy}} & \multicolumn{1}{c|}{\textbf{F1}}    & \multicolumn{1}{c|}{\textbf{Precision}} & \textbf{Recall} \\ \hline
				\multicolumn{1}{|c|}{\cmark}             & \xmark             & \xmark                                                                               & \multicolumn{1}{c|}{87.23}             & \multicolumn{1}{c|}{79.79}          & \multicolumn{1}{c|}{79.56}              & 80.99           & \multicolumn{1}{c|}{70.5}              & \multicolumn{1}{c|}{57.75}          & \multicolumn{1}{c|}{62.25}              & 58.32           \\ \hline
				\multicolumn{1}{|c|}{\xmark}             & \cmark             & \xmark                                                                               & \multicolumn{1}{c|}{94.95}             & \multicolumn{1}{c|}{94.84}          & \multicolumn{1}{c|}{95.67}              & 94.5            & \multicolumn{1}{c|}{92.71}             & \multicolumn{1}{c|}{82.61}          & \multicolumn{1}{c|}{82.53}              & 87.79           \\ \hline
				\multicolumn{1}{|c|}{\xmark}             & \cmark             & Naive Bayes                                                                          & \multicolumn{1}{c|}{96.21}             & \multicolumn{1}{c|}{96.21}          & \multicolumn{1}{c|}{96.4}               & 96.07           & \multicolumn{1}{c|}{94.52}             & \multicolumn{1}{c|}{86.62}          & \multicolumn{1}{c|}{86.91}              & 88.28           \\ \hline
				\multicolumn{1}{|c|}{\cmark}             & \cmark             & Naive Bayes                                                                          & \multicolumn{1}{c|}{96.21}             & \multicolumn{1}{c|}{96.21}          & \multicolumn{1}{c|}{96.4}               & 96.07           & \multicolumn{1}{c|}{95.75}             & \multicolumn{1}{c|}{87.61}          & \multicolumn{1}{c|}{86.8}               & 90.59           \\ \hline
				\multicolumn{1}{|c|}{\xmark}             & \cmark             & GERIS                                                                                & \multicolumn{1}{c|}{96.64}             & \multicolumn{1}{c|}{96.26}          & \multicolumn{1}{c|}{96.67}              & 95.93           & \multicolumn{1}{c|}{96.41}             & \multicolumn{1}{c|}{87.98}          & \multicolumn{1}{c|}{\textbf{89.06}}     & 88.04           \\ \hline
				\multicolumn{1}{|c|}{\cmark}             & \cmark             & GERIS                                                                                & \multicolumn{1}{c|}{\textbf{97.28}}    & \multicolumn{1}{c|}{\textbf{96.79}} & \multicolumn{1}{c|}{\textbf{97.41}}     & \textbf{96.31}  & \multicolumn{1}{c|}{\textbf{97.28}}    & \multicolumn{1}{c|}{\textbf{88.84}} & \multicolumn{1}{c|}{88.49}              & \textbf{91.45}  \\ \hline
			\end{tabular}
		}
	\end{table*}
	
	\section{Discussion}
	
	In this study, the k-nearest neighbor graph is central to the GERIS framework. To find the optimal value for \( h \), we used a grid search strategy. For each candidate value of \( h \), we constructed the k-nearest neighbor graph and ran the GERIS model on the validation set. The value of \( h \) that yielded the highest F1-score on the validation data was chosen as the optimal neighborhood size. This method ensures that the graph structure is adaptively adjusted to maximize the accuracy of noise vector classification.
	
	\subsection{Computational Complexity}
	
	Let \( n \) represent the number of noise vectors and \( d \) the dimensionality of the feature space. The process of building a k-nearest neighbor graph using the brute-force Mahalanobis distance involves computing all pairwise distances between the noise vectors, which has a time complexity of:
	\begin{align}
		O(n^2 d)
	\end{align}
	Since this process is repeated for each candidate value of \( h \) during the grid search (denoted as \( H \) possible values), and assuming the computational cost of label propagation via GERIS is \( T \), the overall time complexity becomes:
	\begin{align}
		O(H \cdot (n^2 d + T))
	\end{align}
	This complexity is manageable for small datasets. In our experiments, the dataset contained a relatively small number of synthetic noise vectors (e.g., 2000 in the first stage and 50000 in the second), which allowed us to use the exact brute-force method for neighbor retrieval without incurring significant computational overhead.
	
	However, scaling this approach to large datasets would result in prohibitive computational costs, especially when graph construction needs to be repeated during the hyperparameter optimization process. To address this challenge, \textbf{Approximate Nearest Neighbor (ANN)} search methods present a promising alternative \cite{wang2021comprehensive}.
	
	\subsection{Future Work: Accelerating Neighbor Search with ANN}
	
	As a direction for future research, we propose replacing the brute-force k-NN graph construction with ANN search methods. These methods aim to identify approximate neighbors for each point, achieving substantially reduced computational cost while sacrificing only a small amount of accuracy. ANN search techniques are particularly useful in high-dimensional or large-scale settings, where exact k-NN search becomes computationally expensive.
	
	ANN algorithms such as Hierarchical Navigable Small Worlds (HNSW), Locality-Sensitive Hashing (LSH), and Product Quantization offer favorable trade-offs between search speed and accuracy. Among available tools, Facebook AI Similarity Search (FAISS) is particularly notable for its performance and ease of use \cite{johnson2019billion}. FAISS provides implementations of several ANN methods and supports GPU-accelerated and CPU-based indexing, batch querying, and memory mapping. It is well-suited for scaling the neighbor search process and can handle large datasets with sub-second query latency.
	
	In the ANN-based extension of GERIS, the framework would be modified as follows:
	\begin{itemize}
		\item The k-nearest neighbor graph is constructed once using ANN methods, and the optimal number of neighbors is determined by search results.
		\item We would select the optimal number of similar records for each query. This eliminates the need for iterative graph construction during hyperparameter optimization.
		\item Owing to the efficiency of ANN search, this extension enables scalable graph construction, making it applicable to larger datasets without sacrificing computational efficiency.
	\end{itemize}
	
	\subsection{Time Complexity of ANN-Based Graph Construction}
	For ANN methods such as HNSW, the time complexity for index construction is approximately:
	$
	O(n \log n)
	$
	and the query time for each data point is sublinear, typically:
	$
	O(\log n)
	$.
	Therefore, constructing a neighborhood graph over all \( n \) points using ANN would have a total time complexity of:
	\begin{align}
		\underbrace{O(n \log n)}_{\text{indexing}}  + \underbrace{O(n \log n)}_{\text{querying}} = O(n \log n)
	\end{align}
	This is a significant improvement over the \( O(n^2 d) \) complexity of the exact k-NN search, particularly when \( n \) is large and \( d \) is moderate to high. By incorporating ANN-based neighbor search into GERIS, we significantly reduce runtime while maintaining the model’s effectiveness in label propagation and noise filtering, making it suitable for deployment on large-scale datasets.
	
	\section{Potential Applications Beyond License Plates}
	
	While our experiments focus primarily on license plate recognition, GERIS’s semi-supervised, graph-pruning approach within the IDN model can easily be applied to other fields:
	
	\begin{itemize}
		\item \textbf{Medical Imaging:} In areas such as histopathology and radiology, subtle variations in texture, illumination, or artifacts can lead to instance-dependent mislabeling (e.g., distinguishing between benign and borderline lesions). By acquiring a small set of confirmed noisy labels—either through expert consensus or known imaging artifacts—GERIS can identify and prune additional mislabeled scans or tiles using a Mahalanobis-distance k-NN graph.
		
		\item \textbf{Remote Sensing:} Satellite imagery frequently suffers from occlusions (such as cloud cover or shadows) that introduce labeling noise in each scene (e.g., misclassifying land-cover types). A limited selection of manually reviewed "noisy" patches can serve as a starting point for GERIS to detect similar mislabeled pixels or segments on a larger scale.
	\end{itemize}
	
	These examples illustrate how GERIS can be effectively utilized under the IDN framework to enhance data quality across various pattern-recognition tasks, especially when a modest number of seed noisy instances is available.

	\section{Conclusion}
	In this study, we addressed the challenge of data augmentation in license plate recognition by introducing GERIS, a game-theoretic framework for filtering synthetic noise vectors.
	GERIS proficiently differentiates between useful and corrupt noise vectors, specifically addressing the critical issue of augmentation-induced noise, which frequently presents itself as instance-dependent label noise (IDN).
	By modeling the instance selection process as a non-cooperative game and using replicator dynamics to reach Nash equilibrium, GERIS effectively classifies noise vectors as either useful or corrupt. Our two-stage empirical evaluation demonstrates that GERIS outperforms classical filtering methods in detecting corrupt instances and leads to improved performance in downstream recognition tasks.
	While this study focused on license plate recognition, the underlying framework of GERIS could be adapted for other domains such as medical imaging, where instance selection is equally critical. These domains, however, may pose additional challenges due to subtler distinctions between noise types and increased difficulty in manual labeling. Exploring such adaptations remains a promising avenue for future research.
	From a scalability perspective, although our current implementation uses exact k-NN search, future work should explore approximate nearest neighbor (ANN) methods to enable efficient deployment on large-scale datasets. Additional improvements may involve refining the payoff structure or incorporating alternative game-theoretic dynamics. Overall, GERIS offers a principled and adaptable approach to improving the quality of augmented datasets across a variety of machine learning applications.

	\bibliographystyle{unsrt}
	\bibliography{BibFile}
	
\end{document}